\documentclass{amsart}

\usepackage{amsmath,amssymb}
\usepackage{amsthm}
\usepackage[alphabetic]{amsrefs}
\usepackage{indentfirst}
\usepackage{mathrsfs}
\usepackage{parskip}
\usepackage{graphicx, wrapfig}
\usepackage{hyperref}
\usepackage{xcolor}
\usepackage{fourier}
\usepackage[margin=1.5in]{geometry}

\usepackage{booktabs}
\usepackage{subcaption, mathtools, hyperref}
\usepackage[export]{adjustbox}

\usepackage{todonotes}

\theoremstyle{plain}
\newtheorem{theorem}{Theorem}
\newtheorem{lemma}{Lemma}

\newtheorem{prop}[theorem]{Proposition}

\theoremstyle{definition}
\newtheorem{defn}[theorem]{Definition}

\theoremstyle{remark}
\newtheorem*{remark}{Remark}

\DeclareMathOperator*{\argmin}{arg\,min}
\DeclareMathOperator*{\argsort}{arg\,sort}

\DeclareMathOperator{\sort}{sort}
\DeclareMathOperator{\sign}{sign}

\newcommand{\RR}{\mathbb{R}}
\newcommand{\NN}{\mathbb{N}}

\IfFileExists{mathabx.sty}%
  {\DeclareFontFamily{U}{mathx}{\hyphenchar\font45}%
   \DeclareFontShape{U}{mathx}{m}{n}{<->mathx10}{}%
   \DeclareSymbolFont{mathx}{U}{mathx}{m}{n}%
   \DeclareFontSubstitution{U}{mathx}{m}{n}%
   \DeclareMathAccent{\widebar}{0}{mathx}{"73}%
}{%
  \PackageWarning{mathabx}{%
    Package mathabx not available, therefore\MessageBreak substituting
    widebar with overline\MessageBreak }%
  \newcommand{\widebar}[1]{\overline{#1}}%
}

\newcommand{\abs}[1]{\left\lvert#1\right\rvert}
\newcommand{\normp}[2]{\left\|#1\right\|_{#2}}

\newcommand{\dap}[1]{\mathrm{d}_{#1_+}}
\newcommand{\dsp}[1]{\mathrm{d}_{#1_{\pm}}}

\title{Bi-Lipschitz Ansatz for Antisymmetric Functions}
\author{Nadav Dym}
\address{Faculty of Mathematics, Faculty of Computer Science, Technion - Israel Institute of Technology, Haifa, Israel}
\email{nadavdym@technion.ac.il}
\author{Jianfeng Lu}
\address{Department of Mathematics, Department of Physics, and Department of Chemistry, Duke University, Durham, NC 27708 USA}
\email{jianfeng@math.duke.edu}
\author{Matan Mizrachi}
\address{Faculty of Mathematics, Technion - Israel Institute of Technology, Haifa, Israel}
\email{mmizrachi@campus.technion.ac.il}
\date{\today}
\begin{document}
\begin{abstract}
Motivated by applications to the simulation of quantum many-body systems by neural networks, researchers have suggested several models which are antisymmetric by construction, and can approximate all antisymmetric functions. However, these works either require very high computational complexity to attain universal approximation, or suffer from discontinuities. In this paper, we introduce two antisymmetric ansatzes which do not suffer from these disadvantages. The first is based on a bi-Lipschitz embedding with respect to a naturally defined metric. The second is a modular anti-symmetrizing projection framework based on the frame-averaging methodology. Both approaches yield continuous antisymmetric models which attain universal approximation guarantees with a polynomial complexity in problem size. Moreover, for both approaches, we obtain quantitative approximation results that bound the number of parameters required to approximate Lipschitz antisymmetric functions to a given accuracy $\epsilon$. We also provide preliminary experimental evidence suggesting improved performance in learning antisymmetric functions.
\end{abstract}

\maketitle

\section{Introduction}

The search for an ansatz for quantum many-body wave functions dates back to the early days of quantum mechanics \cite{slater1929theory}, and has been a central task in quantum chemistry \cite{szabo1996modern}. In recent years, it has received renewed excitement primarily due to advances in neural network-based ansatzes. While it is natural to leverage neural networks' powerful, universal representation to parameterize many-body wave functions, the symmetry requirements of quantum mechanics make this problem distinct from typical neural network applications in deep learning. In particular, for fermionic systems, such as electrons, the Pauli exclusion principle requires that the wave function is totally antisymmetric, i.e., for any $x = \left(x_1, x_2, \ldots, x_n \right) \in \RR^{d \times n}$ and $\sigma \in S_n$, the symmetric group on $n$ letters, we have 
\begin{equation}\label{eq:antisymmetric}
    \Psi(x) = (-1)^{\sigma} \Psi(\sigma x),
\end{equation}
where $(-1)^{\sigma}$ indicates the sign of permutation $\sigma$ and we denote $\sigma x = \left(x_{\sigma(1)}, x_{\sigma(2)}, \ldots, x_{\sigma(n)} \right)$. As a result, a special neural-network-based ansatz that respects the symmetry must be developed. 

Most ansatzes currently used for antisymmetric functions are built upon the backflow ansatz, which dates back to \cite{feynman1956energy}, where the antisymmetry requirement is enforced via a determinant, and each component of the determinant is further parameterized as neural networks. Variational approaches based on such ansatzes have achieved impressive empirical results; see, e.g., \cites{han2019solving,pfau2020ab,hermann2020deep,choo2020fermionic}. 
On the theoretical side, the work \cite{han2019universal} established that such an ansatz is universal; however, their approximation uses a linear combination of an exponential number (in $n$) of determinants to achieve accuracy. Later, it was argued in \cite{hutter2020representing} that only one determinant is enough; the ansatz constructed is discontinuous and hence impractical to use for actual algorithms. Indeed, it was proved in \cite{huang2023geometry} that within the polynomial category, the representation based on the backflow ansatz would require an exponential number of determinants. 
Of course, this does not exclude the possibility of an efficient representation using a less regular ansatz. A more efficient ansatz was recently constructed in \cite{chen2023exact}, with subsequent improvements in \cite{ye2024widetilde}. 
This ansatz is continuous and  involves a combination of symmetric functions, with a polynomial number of antisymmetric Vandermonde determinants of the form 
\begin{equation}\label{eq:vandermonde}
\phi_{y_k}(x)=\prod_{1\leq i<j\leq n} y_k^T(x_i-x_j), \quad k=1,\ldots,m, \end{equation}
where the $y_k, k=1,\ldots,m$ are vectors in $\RR^d$ and $m\geq nd+1$ (the optimal cardinality is actually a bit lower, see \cite{frames}). A disadvantage of this approach is that it is very unstable, as scaling $x$ by $t>0$ leads to scaling of the output of $\phi_{y_k} $ by $t^{n \choose 2} $. While \cite{ye2024widetilde} suggest to handle this problem by straightforward normalization, replacing $\phi_{y_k}(x) $ with $\frac{1}{\sum_{k\in[m]}{\phi_{y_k}(x)}}\sum_{k \in [m]}{\phi_{y_k}(x)} $,
this normalized version of the function  is no longer  continuous.

\subsection{Goals} In this paper, we consider two goals: the first goal is to attain an antisymmetric ansatz with the same low complexity as in \cite{ye2024widetilde}, which yields models which are not only antisymmetric by construction, but are also \textbf{continuous} by construction. Our second goal is to attain \textbf{quantitative approximation rates} for our ansatz: motivated by, and building on, the famous results in \cite{approx_lip} which show that Lipschitz functions can be approximated to $\epsilon$ accuracy on a $D$ dimensional space by a ReLU network with $\sim \epsilon^{-D/2} $ neurons, we will derive similar rates for the approximation of any Lipschitz antisymmetric function by our ansatz.

\subsection{Main Contributions}
In this work, we introduce two different types of antisymmetric ansatz which both fulfill our two goals. 

The first ansatz is based on the development of invariants to the alternating group $A_n$ which are bi-Lipschitz. Applying a ReLU network to these invariant features, together with a simple averaging procedure, we obtain an ansatz which is continuous and antisymmetric by nature, and attains universality with quantitative approximation rates. We also give quantitative bounds for the bi-Lipschitz distortion of our antisymmetric mapping. 

The second ansatz is based on generalized notions of group averaging, introduced in \cites{puny2022frameaveraginginvariantequivariant,frames}. We suggest an averaging technique that averages a ReLU network over (at most) $\sim n^3 $ permutations (rather than the $n!$ permutations needed in full group averaging). We show that the resulting averaged neural networks are continuous and antisymmetric by construction, and again attain universality with quantitative approximation rates. 
    % to formulate an ansatz similar to \cite{chen2023exact}. Compared with the latter, while the ansatzes are similar, our construction is advantageous due to bi-Lipschitzness, which is expected to lead to improved generalization capabilities and more stable learning. Indeed, for symmetric functions, the practical advantages of bi-Lipschitz symmetric models were demonstrated in \cite{fsw,fswgnn,davidson2024h}.
    Finally, we complement our theoretical analysis with  preliminary experiments showing the advantages of our ansatzes over previous work for learning antisymmetric functions. 

\section{Continuous Representation of Antisymmetric Functions}
In this section, we present and discuss our two antisymmetric neural network ansatzes. We begin with the bi-Lipschitz approach. 

\subsection{$A_n$-invariant bi-Lipschitz Embedding}
This approach is based on the close relationship between antisymmetric functions and $A_n$ invariant functions, where $A_n$ denotes the group of permutations with positive sign. Firstly, note that any antisymmetric function is invariant to the action of  $A_n$. Conversely, while not all $A_n$ invariant functions are antisymmetric, this can be alleviated by applying a simple projection procedure. This procedure is defined as follows: fix some permutation $\tau_0$ of negative sign. Then define 
\begin{equation}\label{eq:pi}
	\pi[f](x) =\frac{1}{2}\left( f (x) - f(\tau_0 x) \right).
\end{equation}
It can be easily verified that if $f$ is $A_n$ invariant, then $\pi[f]$ is antisymmetric, and moreover, that if $f$ is already antisymmetric then $\pi[f]=f$. 

Accordingly, our initial focus will be on $A_n$-invariant functions, and the construction of an $A_n$-invariant bi-Lipschitz embedding $\mu$ (to be defined shortly). Once such an embedding is constructed, it is known that we can approximate a general $A_n $ invariant function by $f\circ \mu $ where $f$ is a multilayer perceptron (MLP) or some other neural network architecture. Moreover, recent results by \cite{newPaper} provide quantifiable approximation rates for such bi-Lipschitz constructions. We can then obtain an antisymmetric function by applying the projection $\pi$ from \eqref{eq:pi} to obtain an ansatz $\pi \circ f \circ \mu $ which is both antisymmetric, continuous and enjoys quantifiable approximation guarantees.

To formally define bi-Lipschitz embeddings,  we first define a semi-metric on $\RR^{d\times n}$ by quotienting a standard norm over $A_n$, namely:
\begin{equation}
	\dap{p}{\left(x,y \right)} \coloneq \min_{\sigma \in A_n}{\normp{x-\sigma y}{p}}
\end{equation}
where $p \geq 1$ is some real number, $\|\cdot\|_p$ denotes the elementwise $p$-norm $\|x\|_p^p=\sum_{ij}\abs{x_{ij}}^p $ , and any permutation element $\sigma \in S_n$ is right-acting on any $x \in \RR^{d \times n}$ via  $\sigma x = \left(x_{\sigma(1)}, x_{\sigma(2)}, \ldots, x_{\sigma(n)} \right)$.

%Let us fix some notations before introducing the construction of our ansatz. 
%Fix some $n \geq 2$, denote the symmetric group on $n$ letters by $S_n$ and let $A_n$ denote the alternating group, e.g., the group of permutations $\sigma \in S_n$ whose sign is positive. For $x = (x_1, x_2, \ldots, x_n) \in \RR^{d \times n}$, where $x_i \in \RR^d$ for $i = 1, \ldots, n$, we denote $\sigma x = (x_{\sigma(1)}, x_{\sigma(2)}, \ldots, x_{\sigma(n)})$. 
%On $\RR^{d \times n}$, we define semi-metrics by quotienting the 1-norm over $S_n$ and $A_n$, respectively:
%\begin{align*}
%	d(x,y)&=\min_{\sigma \in S_n} \|x-\sigma y\|_1;\\
%	\da(x,y)&=\min_{\sigma \in A_n} \|x-\sigma y\|_1.
%\end{align*}
Our goal is to define a function $\mu:\RR^{d \times n} \to \RR^m$, for an appropriate dimension $m=m(d,n)$, such that 
\begin{enumerate}
	\item \textbf{Invariance:} $\mu(x)=\mu(\sigma x)$ for all $\sigma \in A_n$ and $x\in \RR^{d \times n}$.
	\item \textbf{Bi-Lipschitz} For all $p \geq 1$, there exist positive $c_1(p),c_2(p)$ such that for all $x,y\in \RR^{d \times n}$,
	\begin{equation}\label{eq:bilip}
	   c_1(p) \dap{p}(x,y)\leq \normp{\mu(x)-\mu(y)}{p} \leq c_2(p) \dap{p}(x,y)  
	\end{equation}
\end{enumerate}
Once such an $A_n$-invariant bi-Lipschitz function $\mu$ is constructed, we can approximate a general $A_n $ invariant function by $f\circ \mu $ where $f$ is a multilayer perceptron (MLP) or some other neural network architecture. We obtain an antisymmetric ansatz by applying the projection $\pi$ from \eqref{eq:pi} to obtain $\pi \circ f \circ \mu $. This ansatz is both antisymmetric and continuous. Compared with \cite{chen2023exact}, although the ansatz is similar, our construction is advantageous due to its bi-Lipschitzness, which is expected to lead to improved generalization capabilities and more stable learning. Indeed, for symmetric functions, the practical advantages of bi-Lipschitz symmetric models were demonstrated in \cites{fsw,fswgnn,davidson2024h}. An additional advantage of antisymmetric ansatz based on bi-Lipschitz models is that they lead to quantitative approximation results:
\begin{prop}
Let $K\subseteq \RR^{d\times n}$ be a compact set which is $S_n$ invariant. Assume that $\mu:K \to \RR^m$ is bi-Lipschitz with bounds $0<c_1 \leq c_2$ for a fixed $p\geq 1$. Then there exist $C=C\left(c_1,c_2, p,n,d,K \right)$ such that for all $\epsilon>0$, and every antisymmetric function $\Psi:\RR^{d\times n} \to \RR$ which is $1$-Lipschitz, there exists  an MLP $f$ with $\leq C\epsilon^{-nd/2} \abs{\log \epsilon} $ parameters, such that 
\begin{equation}
    \max_{x\in K}{\abs{\Psi(x)-\pi\circ f \circ \mu(x)}}<\epsilon. 
\end{equation}
\end{prop}
\begin{proof}
Theorem 21 in \cite{newPaper} gives quantifiable approximation rates for invariant Lipschitz functions using models of the form $f\circ \mu$ where $f$ is an MLP and $\mu$ is bi-Lipschitz. Namely,  for some constant $C=C\left(c_1,c_2, p,n,d,K \right)$, the theorem guarantees   that for all $\epsilon>0$ and any  $\Psi: \RR^{d\times n} \to \RR$ that is $1$-Lipschitz and antisymmetric, there exists a ReLU network $f$ with $\leq C\epsilon^{-nd/2} \abs{\log \epsilon} $ parameters such that 
\begin{equation}
    |\Psi(x)- f\circ \mu(x)|\leq \epsilon, \quad \forall x\in K .
\end{equation}
It remains only to show that this approximation is maintained under projection by $\pi$. Indeed
\begin{equation}
    \begin{split}
     {\abs{\Psi(x)-\pi\circ f \circ \mu (x)}}&= {\abs{\pi \circ\Psi(x)-\pi\circ f \circ \mu (x)}}\\
     &=\frac{1}{2}|\Psi(x)-\Psi(\tau_0x)-f\circ \mu(x)+f\circ \mu(\tau_0 x) |\\
&\leq \frac{1}{2}|\Psi(x)-f\circ \mu(x)|+\frac{1}{2}|\Psi(\tau_0x)-f\circ \mu(\tau_0 x) |\\
&\leq \frac{\epsilon}{2}+\frac{\epsilon}{2}   \qedhere
    \end{split} 
\end{equation}
\end{proof}

\begin{remark}
We note that Theorem 21 in \cite{newPaper} is more general: firstly, it discusses the case where the domain $K$ has low intrinsic dimension, in the sense that it can be covered by $\sim \epsilon^{-\nu}$ balls, where $\nu \leq nd$. It also discusses approximations of $\Psi$ that are $\alpha$-Holder and not only the Lipschitz case $\alpha=1$. In these settings, the network size will be $\sim \epsilon^{-\nu/2\alpha} $. Accordingly, our approximation result above could be extended to these cases as well. 

Additionally, we note that for any \emph{continuous} $\Psi$, we can find a ReLU network $f$ (of unknown size) which $\epsilon$ approximates it, since bi-Lipschitz $\mu$ is in particular injective (see \cite{dym2024low}). The Lipschitz assumption on $\Psi$ is only necessary to obtain \emph{quantifiable} approximation rates. 
%, as this paper focuses primarily on finding a bi-Lipschitz construction.   
%Moreover, if $K$ resides in some linear subspaces of $\RR^{d\times n}$ of strictly lower dimension $\Delta $, the bound can be improved to $2\Delta+1$. This will be discussed in XXX. 
\end{remark}
We now give the construction of our bi-Lipschitz embedding. We will in fact consider two variants: we will first consider a bi-Lipschitz invariant $\xi$ for which we can also quantify the bi-Lipschitz constants. Then we will apply random linear projections to $\xi$ to obtain a more efficient bi-Lipschitz invariant $\mu$ (for which we do not quantify the bi-Lipschitz constants). We begin with constructing $\xi$ in the case where $d=1$:
% First, we will construct two of the three components of our embedding in a simple setting of $d=1$ to present the key ideas. We then generalize their construction to $d>1$, and address an initial $A_n$-invariant bi-Lipschitz embedding $\xi$ for which we quantify the bi-Lipschitz constants with respect to all convex $p$-norms. Finally, we construct the third component $\delta$, which we compose on $\xi$ to define $\mu$, an $A_n$-invariant bi-Lipschitz embedding with reduced complexity. Still, we do not specify estimates for the latter's bi-Lipschitz constants, as they require work outside our scope.
\subsubsection{Construction of bi-Lipschitz Embedding} \label{sssec:bl1}
We first focus on the case $d=1$. We begin the construction with a \emph{symmetric} bi-Lipschitz function. That is, a symmetric function which is bi-Lipschitz with respect to the metric obtained by quotienting over the group $S_n$ of all permutations, namely
\begin{equation} \label{eq:ds} \dsp{p}(x,y) \coloneq \min_{\sigma\in S_n}{\normp{x-\sigma y}{p}}, \quad p \geq 1,
\end{equation}
As shown in \cite{amir}, commonly used symmetric models based on summation, as suggested in DeepSets \cite{zaheer2017deep}, are not bi-Lipschitz. In contrast, following \cite{balan2022,davidson2024h} we choose the symmetric function $\sort:\RR^n \to \RR^n $ that maps each vector $x$ to the unique vector $\hat{x}=\sort(x) $ satisfying that $\hat x_i \leq \hat x_{i+1} $ for all $i=1,\ldots,n-1$.\footnote{The idea of sorting has also been used in the antisymmetric construction in \cite{hutter2020representing} in one dimension.} The $\sort$ function is continuous and piecewise linear. It is not only bi-Lipschitz, but it is also an \emph{isometry}, namely 
\begin{equation}
    \normp{\sort{(x)}-\sort{(y)}}{p}= \dsp{p}(x,y). 
\end{equation}
We note that while $\hat{x}$ is uniquely defined, the permutation taking $x$ to $\hat x$ is not uniquely defined if $x$ is in the set 
\begin{equation} \label{def:omega}
    \Omega_n :=\bigl\{x\in \RR^{n} \;\mid \; x_i=x_j \text{ for some } 1\leq i<j\leq n\bigr\}
\end{equation}
In this case, there will be permutations of both positive and negative signs that map $x$ to $\hat{x}$. However, if $x$ is not in $\Omega_n$, the permutation sorting $x$ is unique, and we denote it by $\tau_x$. 

The second component of our construction is the function $Q:\RR^n \to \RR$ defined as
\begin{equation} \label{def:Q}
    Q(x)=\left(\prod_{1\leq i<j \leq n} \sign(x_j-x_i) \right) \cdot \min_{1\leq i<j\leq n}{\abs{x_j-x_i}}
\end{equation}
This function is continuous, piecewise linear, and antisymmetric. Moreover, we note that $Q(x)=0$ if and only if $x \in \Omega_n$. Next, we note that if $x\not \in \Omega $  and $\hat x=\sort(x)$, then the sign of $Q(\hat x)$ will be positive. These observations give us an efficient method for computing $Q$: for every $x\in \RR^n$, if $x$ is in $\Omega_n$ then $Q(x)=0$. Otherwise, there is a unique permutation, which we denote as before by $\tau_x$, such that 
$\hat x=\sort(x)=\tau_x x $. We then deduce that
\begin{equation}
    Q(x)=(-1)^{\tau_x}Q(\tau_x x)=(-1)^{\tau_x}\min_{1\leq i<j\leq n} |\hat x_j-\hat x_i|=(-1)^{\tau_x}\min_{1\leq j\leq n} \bigl( \hat x_{j+1}-\hat x_j\bigr). 
\end{equation}
It follows that $Q(x)$ can be computed by first sorting $x$ using $\mathcal{O}(n\log n)$ operations, and then applying the formula above, which requires only $\mathcal{O}(n)$ operations. We now combine the two functions to define 
\begin{equation}
    \xi(x) = \begin{bmatrix}
\sort{(x)} \\ Q(x)
\end{bmatrix}.
\end{equation}
The function $\xi$ is antisymmetric. It is injective and bi-Lipschitz, as we will soon prove in Theorem \ref{thm:bl1} for the more general setting of $d \geq 1$. What is special about the $d=1$ setting is that we can obtain bi-Lipschitz bounds $c_1,c_2$ which are reasonably small and do not depend on $n$. For example for $p=1$ we obtain $c_1=1,c_2=2$. The bounds for general $p\geq 1$ are stated and proved in \href{sec:appendixA}{Appendix A}.

We extend the definition of $\xi$ to the  $d > 1$ setting by applying  $m$ linear projections and then applying the $d=1$ version of $\xi$ to each projection. Specifically, we consider the parametric functions $\beta_A: \RR^{d \times n} \to \RR^{m \times n}$ and $\bar{Q}_A: \RR^{d \times n} \to \RR^{m}$, parameterized by a matrix $A \in \RR^{m \times d}$, and defined by
\begin{equation} \label{def:betaQ}
\begin{split}
\beta_A(x) &= \begin{bmatrix}
\sort{\left(x^T a_k \right)}
\end{bmatrix}^T_{k \in [m]}, \\
\bar{Q}_A(x) &= \begin{bmatrix}
Q\left(x^T a_k \right)
\end{bmatrix}_{k \in [m]}
\end{split}
\end{equation}
where $a_k$ are the rows of $A$. For $m=n(d-1)+1$ and generic $A$, these two mappings form a bi-Lipschitz embedding:
\begin{theorem} \label{thm:bl1}
    Set $m = n(d-1) + 1$. Then, for Lebesgue almost every $A \in \RR^{m \times d}$, the $A_n$-invariant embedding
    \begin{equation} \label{eq:xi}
    \xi(x ; A) \coloneq \begin{bmatrix}
    \beta_A(x) & \bar{Q}_A(x)
\end{bmatrix}
    \end{equation}
is injective on the quotient space $\RR^{d \times n} / A_n$, and bi-Lipschitz with respect to $\dap{p}$ for all $p\geq 1$.
\end{theorem}
\begin{proof}
We begin by showing injectivity. This relies on two previous results: according to Theorem 3 in \cite{dym2025quantitativeboundssortingbasedpermutationinvariant}, for $m = n(d-1) + 1$, Lebesgue almost every $A \in \RR^{m \times d}$ defines $\beta_A$ injectively on the quotient space $\RR^{d \times n} / S_n$. According to \cite{frames}, for almost every $A  \in \RR^{m \times d}$, we will have for every $x \in \RR^{d\times n} \setminus \Omega_{d \times n}$ that for some $k$, the vector $x^Ta_k $ will not have repeated entries, or equivalently, $Q(x^Ta_k)\neq 0 $. It follows that for generic $A$, the combined mapping $\xi(\cdot;A)$ is injective
 on the quotient space $\RR^{d \times n} / A_n$. To see this, consider two distinct matrices $x, y \in \RR^{d \times n}$ which are not related by a positive signed permutation. If $x, y$ are also unrelated by a negative sign permutation, then $\beta_A(x) \ne \beta_A(y)$. Alternatively, if $x = \sigma y$ for some negative signed $\sigma \in S_n$ but not for any positive signed permutation, then in particular $x, y$ are not in $\Omega_{d \times n}$, and therefore there exists $k \in [m]$, such that $0 \neq Q\left(x^T a_k \right) = -Q\left(y^T a_k \right)$ and so the injectivity of $\xi$ is concluded. 
 
 We now show that the embedding is bi-Lipschitz. Owing to its injectivity on $\RR^{d \times n} / A_n$, the embedding satisfies $\normp{\xi(x ; A) - \xi(y ; A)}{1} = 0$ if and only if $\dap{p}(x, y) = 0$, for all $x, y \in \RR^{d \times n}$. Applying this for $p=1$, we obtain two piecewise linear non-negative functions with the same zero set, and by  Lemma 3.4 in \cite{fswgnn} they are bi-Lipschitz equivalent on the polyhedral compact set $\{(x, y) \mid  \|x\|_1+\|y\|_1\leq 1\}$. The homogeneity of $\xi$ and the  $1$-norm implies that the bi-Lipschitz equivalence in fact holds on the whole domain. Finally, the equivalence of norms implies bi-Lipschitzness for any choice of  $p$.
\end{proof}
In \href{sec:appendixA}{Appendix A} we provide quantification for the Lipschitz bounds of $\xi$, with respect to $\dap{p}$. 

\subsubsection{More efficient bi-Lipschitz embedding} \label{sssec:bl2}
The mapping $\xi$ maps a given $x\in \RR^{d\times n}$ to a matrix of dimension $m\times (n+1) $. Since the $m $ required by Theorem \ref{thm:bl1} grows linearly in $n$, the overall embedding dimension $m(n+1) $ grows quadratically in $n$. We now obtain a more efficient bi-Lipschitz embedding $\mu$ by applying linear projections to $\xi$.  For all $1 < d, n \in \NN$, and an appropriately chosen $m \in \NN$, we consider a matrix $B \in \RR^{m \times (n+1)}$ and the parametric function $\eta_B: \RR^{m \times (n+1)} \to \RR^{m}$, defined by
\begin{equation}
    \eta_B(x) = \begin{bmatrix} b_i \cdot x_i \end{bmatrix}_{i \in [m]}
\end{equation}
where the notation $x\cdot y$ denotes the standard inner product between $x$ and $y$ in $\RR^{n+1}$.
\begin{theorem}\label{thm:multi}
Set $m=2nd+1$.
For Lebesgue almost every $A \in \RR^{m \times d}$ and $B \in \RR^{m \times (n+1)}$, the function 
\begin{equation}
    \mu(x; A, B)=  \eta_B \circ \xi(x ; A) = \begin{bmatrix} b_{k, 1:n} \cdot \sort{\left(x^T a_k \right)} + b_{k, n+1}Q\left(x^T a_k \right)  \end{bmatrix}_{k \in [m]} 
\end{equation}
is one-to-one (up to $A_n$ symmetries) and bi-Lipschitz.
\end{theorem}
\begin{proof}
We first show injectivity, up to $A_n$ symmetries. In Theorem \ref{thm:bl1}, we use $m=n(d-1) + 1$ to prove the injectivity, up to $A_n$ symmetries, of $\xi(\cdot; A)$ for Lebesgue almost every $A \in \RR^{m \times d}$. That is, for all $x, y \in \RR^{d \times n}$ which are not related by an $A_n$ symmetry, and for Lebesgue almost every collection $A = \left\{a_k \right\}_{k \in [2nd+1]} \subset \RR^{d}$, there exists $a \in A$ such that $\xi(x; a) \ne \xi(y; a)$. Hence, for the any $a \in \RR^d$ fulfilling the latter inequality, we note that for Lebesgue almost every collection $B = \left\{b_k \right\}_{k \in [2nd+1]} \subset \RR^{n+1}$, there exists $b \in B$, such that $b \cdot \xi(x; a) \ne b\cdot \xi(y; a)$. We will now use Theorem 1.7 in \cite{dym2024low}, which guarantees the injectivity (separation in the language of that paper) of $\mu(\cdot; A, B)$ using $m=2nd+1$ generic vectors; consequently, to the latest inequality, up to $A_n$ symmetries.
Next, we prove bi-Lipschitzness. For this, we will again use Lemma 3.4 from \cite{fswgnn}. This lemma considers two piecewise linear, non-negative functions that share the same zero set. In our case, these functions are 
\begin{equation}
    \dap{p}(x,y), \quad p=1 \quad \text{ and } \quad \| \mu(x;A, B)-\mu(y;A, B)\|_1
\end{equation}
where the parameter $A$ is such that $\xi(\cdot; A)$ is injective, so that the two functions have the same zero set. 
The lemma says that these two functions will be bi-Lipschitz equivalent on a fixed compact polyhedral domain, which in our case we will take to be the unit ball $\{(x,y)\mid \|x\|_1+\|y\|_1\leq 1 \} $. Homogeneity of $\dap{p}$ and $\mu$ then implies bi-Lipschitzness on the whole domain, and the equivalence of norms implies the result for $p>1$.  
\end{proof}
\begin{remark}
The cardinality of $m$ chosen here can probably be improved a bit, e.g., using the scale and translation invariance of the question at hand. Moreover, if we assume that $x,y$ come from some semialgebraic subset of $\RR^{d\times n}$ of dimension $\nu$, then we can get a bound of $m=2\nu+1$ essentially automatically. 
\end{remark}

\section{Antisymmetric Frame Averaging Networks} 
\label{ssec:afanet}
We now introduce our second method for attaining continuous antisymmetric models with quantifiable approximation guarantees. This method is based on frame averaging methodology, which we will now review. 

An antisymmetric function is a special case of an equivariant function: in the general setting, we consider a group $G$, two linear spaces $V_1,V_2$ on which $G$ acts linearly via $\rho_1$ and $\rho_2$ respectively. An equivariant function is a function $f:V_1\to V_2$ satisfying that $f(\rho_1(g)x)=\rho_2(g)f(x) $ for all $g\in G$ and $x\in V_1$. An antisymmetric function is an equivariant function in the special case where the group considered is $G=S_n$, which acts on the space $V_1=\RR^{d\times n}$ by permuting columns, and on  $V_2=\RR$ by multiplication by the sign of the permutation.

 A classical way to construct equivariant functions, from arbitrary base functions, is to average a function over the action of the symmetry group. Namely, by applying the following operator:
\begin{equation}
    \mathcal{E}[f](x) =
\frac{1}{\abs{G}}
\sum_{g \in G}
\rho_2(g) \cdot f(\rho_1(g)^{-1} x).
\end{equation}
However, computing the full group average can be computationally expensive when the group is large,  as in our case, where we consider the permutation group. To address this limitation, Puny et al.~\cite{puny2022frameaveraginginvariantequivariant} introduce the concept of \emph{frame averaging}. Instead of averaging over the entire group, their method averages over a subset of group elements called a frame. In this framework, a frame is defined as a set-valued function $\mathcal{F} : V \rightarrow 2^{G} \setminus \emptyset$ which is equivariant, in the sense that $\mathcal{F}(\rho_1(g)x)=g\mathcal{F}(x) $.  Using such a frame, an equivariant projection operator is defined via
\begin{equation}
\mathcal{E}[f](x)
=
\frac{1}{\abs{\mathcal{F}(x)}}
\sum_{g \in \mathcal{F}(x)}
{\rho_2(g)\cdot f(\rho_1(g)^{-1}x)}. 
\end{equation}
As shown in \cite{puny2022frameaveraginginvariantequivariant}, this construction is guaranteed to produce equivariant functions, while significantly reducing the computational cost compared to full group averaging. However, these frames often do not preserve continuity, meaning that, even if the function $f$ is continuous, the function $ \mathcal{E}[f]$ may not be. 
Subsequent theoretical work \cite{frames}  shows that for functions on $\RR^{d\times n}$ which are \emph{permutation invariant} (that is, $\rho_2$ is trivial), only the full frame which contains all group elements (leading to group averaging) can preserve continuity. On the positive side, they show that continuity preservation with frames of moderate size can be obtained using weighted frames,  where instead of averaging uniformly over the elements of the frame, the averaging is performed using weights assigned to the elements of the frame. They provide an explicit construction of a continuity preserving weighted frame for permutation invariant functions which preserves continuity, and whose size depends linearly on $n,d$. However, this frame only preserves continuity outside of $\Omega_{d\times n}$. Our goal will be to show how their frame can be extended to a continuity preserving frame on all of $\RR^{d\times n}$, in the equivariant setting of antisymmetric functions. 

Our starting point is thus a weighted permutation frame, such as the frame provided by \cite{frames} (this specific frame will be reviewed later on), which has the following properties:

\begin{defn}[Weighed Permutation Frame]\label{def:WPF}
Let $d,n,m$ be natural numbers. Denote the $m$-dimensional simplex by $\Delta_m$. A weighted permutation frame of size $m$ is a function $\mathcal{F}_{\text{as}}: \RR^{d \times n} \setminus{\Omega_{d \times n}} \to \left(S_n \right)^m \times \Delta_m$, denoted by $\mathcal{F}_{\text{as}}(x)=(\tau_i,w_i)_{i=1}^m $, such that 
\begin{enumerate}
\item  For all $x \in \RR^{d \times n}\setminus{\Omega_{d \times n}}$, all $ \sigma \in S_n$, and all $i \in [m]$,
    \begin{equation} \label{eq:frame_cond}
       \tau_i\left( \sigma \cdot x \right) = \sigma \tau_i(x), \quad
       w_i(\sigma \cdot x) = w_i(x).
\end{equation}
\item The mapping $x \mapsto \sum_{i=1}^m w_i(x)\delta_{\tau_i(x)} $ is a continuous mapping from $\RR^{d\times n}\setminus \Omega_{d\times n}$ to the space of Borel Measures on $S_n$ (with respect to the weak topology).
\end{enumerate}
\end{defn}

As shown in \cite{frames}, these two conditions imply that applying the averaging operation 
\begin{equation}\label{eq:pre}
\mathcal{E}[f](x) =\sum_{i=1}^m w_i(x)(-1)^{\tau_i}f\left( \tau_i^{-1} x \right)\end{equation}
to a continuous function $f$ on $\RR^{d\times n} \setminus \Omega_{d\times n} $, will give a continuous and antisymmetric function on $\RR^{d\times n} \setminus \Omega_{d\times n} $. the challenge is extending this to all of $\RR^{d\times n}$. We achieve this by utilizing the fact that the value of antisymmetric functions on $\Omega_{d\times n} $ is always zero, and so augmenting the averaging operation in \eqref{eq:pre} with an additional operator $\Phi$ which guarantees a continuous transition to zero as points approach $\Omega$. This operator is required to have the following properties:

\begin{defn}[WS and Antisymmetric Frame Averaging Operator] \label{def:afa}
Let $d,n,m$ be natural numbers, and let $\mathcal{F}_{\text{as}}$ be a weighted permutation frame of size $m$. Let  $\Phi[\cdot]$ be a linear operator mapping functions on $\RR^{d\times n}\setminus \Omega_{d\times n}$ to functions on the same space. We say that $\Phi $ is a \textit{weakly stabilizing (WS) operator} for $\mathcal{F}_{\text{as}}$, if
\begin{enumerate}

\item For every compact $K \subset \RR^{d \times n}$, which is $S_n$-invariant, there exists a positive constant $C_K$, such that
  \begin{equation} \label{eq:cond_bound_antis}
    \forall f\in  C\left(\RR^{d \times n}\right), \quad   \max_{x \in K}{\abs{\Phi[f](x)}} \leq C_K \max_{x \in K}{\abs{f(x)}}
  \end{equation}
  \item If $f:\RR^{d\times n} \to \RR$ is antisymmetric, then $\Phi[f] = f$.
    \item \label{cond:proj} For every $f\in  C\left(\RR^{d \times n} \setminus \Omega_{d\times n} \right)$, the function $\Phi[f]$ will also be continuous on $\RR^{d\times n}\setminus \Omega_{d\times n} $.
    \item For any convergent sequence 
  $x^k \to x$ in $\RR^{d \times n} \setminus{\Omega_{d \times n}}$, with limit $x \in \Omega_{d \times n}$,
  \begin{equation} \label{eq:cond_weak_convergence}
\forall f\in  C\left(\RR^{d \times n}\right), \quad     \Phi[f]\left(x^k \right)  \stackrel{k \to \infty}{\rightarrow} 0
  \end{equation}

\end{enumerate}
An \emph{antisymmetric frame averaging operator} will then be an operator
\begin{equation} \label{eq:form2frame}
    \begin{split}
        \mathcal{E}[f](x) &\coloneq 0, \quad \forall x \in \Omega_{d \times n}, \\
        \mathcal{E}[f](x) &\coloneq \sum_{i=1}^{m}{w_i(x)(-1)^{\tau_i}\Phi[f]\left( \tau_i^{-1} x \right)}, \quad \forall x \in \RR^{d \times n} \setminus{\Omega_{d \times n}}.
    \end{split}
    \end{equation}
    defined by a weighted permutation frame $\mathcal{F}_{\text{as}}$ and an appropriate WS operator $\Phi$. 
\end{defn}
% \begin{remark}
%   For a frame of fixed cardinality $m \in \NN$, the projection \eqref{eq:form2frame} could be further generalized, so that $m$ distinct operators $\Phi_i[\cdot]$ would be applied on $f$.
% \end{remark}
An antisymmetric frame averaging operator, is defined above, defines a continuity preserving projection onto antisymmetric functions, as the following lemma shows: 
\begin{lemma} \label{lemma:bec}
Let $d,n$ be natural numbers, and let $K\subseteq \RR^{d\times n}$ be an $S_n$ invariant compact set. An antisymmetric frame averaging operator $\mathcal{E}[\cdot]$ is a bounded linear projection from $C(K) $ to the space of antisymmetric functions in  $C(K)$.
\end{lemma}
\begin{proof}
  We need to prove several properties:

 \textbf{Anti-symmetric projection:} We will show that for any $f\in C(\RR^{d\times n})$, the function $\mathcal{E}[f] $ is antisymmetric. Indeed, Let $x \in \RR^{d \times n}$. By definition, if $x \in \Omega_{d \times n}$, then $\mathcal{E}[f](x) = 0$. Otherwise, if $x \in \RR^{d\times n} \setminus{\Omega_{d \times n}}$, then for all $\sigma \in S_n$, we have
  \begin{equation} \label{eq:antisframecond1}
\begin{split}
  \mathcal{E}[f](\sigma x) &= \sum_{\left(\tilde{\tau}_i, w_i \right) \in \mathcal{F}_{\text{as}}(\sigma x)}{w_i(\sigma x)(-1)^{\tilde{\tau}_i}\Phi[f]\left(\tilde{\tau}_i^{-1} \sigma x \right)} \\ 
  &\stackrel{\tilde{\tau}_i = \sigma \tau_i}{{=\joinrel=\joinrel=}} \sum_{\left(\sigma \tau_i, w_i \right) \in \mathcal{F}_{\text{as}}(x)}{w_i(x)(-1)^{\tau_i}(-1)^{\sigma}\Phi[f]\left(\tau_i^{-1} \sigma^{-1}\sigma x \right)} = (-1)^{\sigma}\mathcal{E}[f](x)
\end{split} 
\end{equation} 
as desired. Next, if $f$ is antisymmetric, then by the second condition in definition \ref{def:afa} $\Phi[f]=f$, and  so  we have
\begin{equation} \label{eq:antisframecond2}
\begin{split}
  \mathcal{E}[f](x) &= \sum_{i=1}^{m}{w_i(x)(-1)^{\tau_i}\Phi[f]\left(\tau_i^{-1} x \right)} \\ 
  &= \sum_{i=1}^{m}{w_i(x)(-1)^{\tau_i}\cdot (-1)^{\tau_i}f\left( x \right)} = f(x),
\end{split} 
\end{equation}
hence, we conclude that $\mathcal{E}[\cdot]$ is an antisymmetric projection.

\textbf{Continuity preservation} Next, we need to show that if $f\in C(K)$ then $\mathcal{E}[f] $ will also be continuous. The continuity of $\mathcal{E}[f]$ within $\RR^{d \times n} \setminus{\Omega_{d \times n}}$ is given by the continuity of $\mathcal{F}_{\text{as}}$ and $\Phi[f]$ in this domain, Equation \eqref{eq:cond_weak_convergence}  ensures the continuity of $\mathcal{E}[f]$ at $\Omega_{d \times n}$, and thus $\mathcal{E}[f]$ is continuous everywhere. 

\textbf{Linearity} follows easily from the fact that $\Phi$ is linear by assumption. 

\textbf{Boundedness} $\mathcal{E}$ is a bounded operator on $C(K)$, as for any $f\in C(K) $,
  \begin{equation} \label{eq:lemma_bec_uni}
    \begin{split}
       \max_{x \in K}{\abs{\mathcal{E}[f](x)}} 
       &= \max_{x \in K}{\abs{\sum_{i=1}^{m}{w_i(x)(-1)^{\tau_i}\Phi[f]\left( \tau_i^{-1} x \right)}}}  \\
    &\leq 
    \max_{x \in K}{\sum_{i=1}^{m}{w_i(x) \abs{\Phi[f]\left( \tau_i^{-1} x \right)}}}  \\ 
 &\leq       \max_{x \in K}{\abs{\Phi[f]\left(x \right)}}
    \\
    &\stackrel{\text{Eq. } \eqref{eq:cond_bound_antis}}{\leq}  C_K \max_{x \in K}{\abs{f(x)}}   
        \end{split}
    \end{equation}
  to conclude the boundedness of $\mathcal{E}[\cdot]$. With that, we have shown that $\mathcal{E}[\cdot]$ defined in definition \ref{def:afa} is a bounded, antisymmetric and continuity preserving linear projection.
\end{proof}
% \begin{defn}[Stability under antisymmetry] \label{def:stable_func}
%   A function $f: \RR^{d \times n} \to \RR$ is \textit{stable} under antisymmetry, if it satisfies $\stab_{S_n}(x) \subseteq \stab_{S_n}\left(f \left( x\right) \right)$ for all $x \in \RR^{d \times n}$. That is, 
%   \begin{equation} \label{eq:stab}
%   \left \{\sigma \in S_n : \sigma \cdot x =x \right\} \subseteq \left \{\sigma \in S_n : (-1)^{\sigma} f(x) = f(x) \right\}, \quad \forall x \in \RR^{d \times n}
% \end{equation}
% or, equivalently, $f(\Omega_{d \times n}) \equiv 0$, with $\Omega_{d \times n}$ defined at \eqref{def:omega}.
% \end{defn}
The positive properties of antisymmetric frame averaging, as proven in lemma \ref{lemma:bec}, lead to an antisymmetric model with quantifiable approximation guarantees:
\begin{theorem}[Universality of Linear Antisymmetric Weighted Frame Averaging] \label{theorem:universality_asc}
Suppose $\mathcal{E}[\cdot]$ is an antisymmetric frame averaging operator, as defined in Definition \ref{def:afa}. Then, for compact $K \subseteq [0,1]^{d \times n}$, which is $S_n$-invariant, there exists positive constant $C = C(n, d, K)$, such that for all $\varepsilon > 0$ and any 1-Lipschitz antisymmetric function $\Psi: \RR^{d \times n} \to \RR$, there exists an MLP $f: \RR^{d \times n} \to \RR$ of $\sim C \varepsilon^{-\frac{1}{2}n \cdot d}$ parameters, so that 
\begin{equation}\label{eq:approx}
\max_{x\in K} |\Psi(x)- \mathcal{E}[f](x)|<\varepsilon. \end{equation}
\end{theorem}
\begin{proof}
  Let $\Psi: \RR^{d \times n} \rightarrow \RR$ be any 1-Lipschitz antisymmetric function. In lemma \ref{lemma:bec}, we've shown $\mathcal{E}[\cdot]$ defines an antisymmetric linear projection operator, which is bounded as in \eqref{eq:lemma_bec_uni}, with some positive constant $\tilde{C} = \tilde{C}(K)$. And so, for all $f: \RR^{d \times n} \to \RR$, we have \begin{equation}\label{eq:bounded_universality}
  \begin{split}
          \max_{x \in K}{\abs{\mathcal{E}[f](x) - \Psi(x)}} &= \max_{x \in K}{\abs{\mathcal{E}[f - \Psi](x)}} \\
          &\leq \tilde{C} \max_{x \in K}{\abs{f(x) - \Psi(x)}} .
  \end{split}
  \end{equation}
  The function $\Psi$ can be extended to a 1-Lipschitz function on all of $[0,1]^{d\times n}$ by the Kirszbraun theorem, and so,
  according to \cite{yarotsky2018optimal}, there exists a constant $C = C(n, d)$, such that for all $\varepsilon > 0$, there exists an MLP $f: [0, 1]^{d \times n} \to \RR$ of $\sim C \varepsilon^{-\frac{1}{2}n \cdot d}$ parameters, such that
  \begin{equation} \label{eq:pre_bounded_universality}
      \tilde{C} \max_{x \in K}{\abs{f(x) - \Psi(x)}} < \varepsilon
  \end{equation}
 combining this with equation \eqref{eq:bounded_universality}, gives us the required inequality in \eqref{eq:approx}. \end{proof}

\begin{remark}
  Theorem \ref{theorem:universality_asc} will holds for any  $S_n$-invariant compact set $K \subset \RR^{d \times n}$ (with different constants), even if it is not contained in $[0,1]^{d\times n}$  This can be seen by applying a linear scaling function to map $K$ into $[0,1]^{d\times n} $.
\end{remark}
\subsubsection{Construction of an Antisymmetric Frame Averaging Operator} \label{sec:construction_afat}
We now give an explicit construction of an antisymmetric frame averaging operator $\mathcal{E}[\cdot]$. Recall this is defined via a weighted permutation frame $\mathcal{F}_{\text{as}}$ and an appropriate WS operator $\Phi$. 

We use the weighted permutation frame introduced in \cite{frames} (see section 4.1); this frame is based on a choice of $m$ vectors $a_1,\ldots,a_m \in \RR^d $. Each vector represents a direction on which to project a given input $x$, and it is assigned a corresponding permutation which sorts $x$ according to this projection, namely: 
\begin{equation} \label{eq:tau}
	 \tau_i^{-1}(x) \coloneq \argsort{\left(x_1 \cdot a_i, \dots, x_n \cdot a_i \right)}  .\end{equation}
We note that if two points $x_k,x_\ell$ have the same projection onto $a_i$, then $\tau_i$ is not uniquely defined. However, it is shown in \cite{frames} that every generic choice of $m\geq n(d-1) $ vectors $a_i$ gives a \emph{globally separating family}, which is equivalent to saying that for every $x\in \RR^{d\times n} \setminus \Omega_{d\times n}$, at least one of the permutations $\tau_i$ is well defined. To avoid permutations which are not well defined, the weights assigned to each $\tau_i$ are constructed to vanish when this issue occurs. These weights are  defined via
\begin{equation}\label{eq:w}
	w_i(x) \coloneq \frac{\abs{Q\left(x \cdot a_i \right)}}{\sum_{l \in [m]}{\abs{Q\left(x \cdot a_l \right)}}}=\frac{\min_{1\leq j<k\leq n} |a_i \cdot(x_j-x_k) |  }{\sum_{l\in[m]}\min_{1\leq j<k\leq n} |a_l \cdot(x_j-x_k) | },
\end{equation}
and the fact that $a_1,\ldots,a_m$ are globally separating implies that there is no division by zero in the definition of $w_i(x)$, for all $x \in \RR^{d\times n} \Omega_{d\times n}$. 

As shown in \cite{frames}, this construction forms a weighted permutation averaging frame, in the sense of Definition \ref{def:WPF}. 
%
%here $\argsort: \RR^n \to S_n$ could be considered as well-defined everywhere, that is
%\begin{equation} \label{eq:argsortdef}
%  \forall x \in \RR^n \text{ if } x_i = x_j \text{ and } i < j \implies \argsort(x)(i) < \argsort(x)(j)
%\end{equation}
%and $Q: \RR \to \RR$ defined at \eqref{def:Q}.
 We now construct a WS operator $\Phi[\cdot]$. 
Let 
\begin{equation}\label{eq:all_transpositions}
    H \coloneq \left \{
    \sigma_{i,j} = \begin{pmatrix}
  i, j
\end{pmatrix} \in S_n  \right\}
\end{equation}
be the subset of all transpositions in $S_n$, and $\eta(x) \coloneq \min_{i < j}{\normp{x_i - x_j}{2}}$.
The operator $\Phi$ maps any $f: \RR^{d \times n} \to \RR$ to a new function $\Phi[f]$ defined by
% \begin{equation} \label{def:staboperator}
%     \Phi[f](x) \coloneq \sign{\left(f(x) \right)}\sqrt{\frac{1}{2}\abs{f(x)}\min_{\sigma \in H}{\abs{f\left( X \right) - f\left( \sigma X \right) }} }, \quad X \in \RR^{d \times n}
% \end{equation}
\begin{align} \label{def:staboperator}
        \Phi[f](x) &\coloneq f(x) - \sum_{\sigma_{i,j} \in H}{\omega_{i,j}(x)\left(f(x) + f(\sigma_{i,j} x) \right)}, \quad \forall x\in \RR^{d\times n}\setminus \Omega_{d\times n}\\
         \Phi[f](x) &\coloneq 0, \quad \forall x \in \Omega_{d\times n} \nonumber
\end{align}
 where $\omega_{i,j}$ are defined via:
\begin{equation} \label{def:small_omega} \begin{split}
\tilde{\omega}_{i,j}(x) &\coloneq \cot \left( \frac{\pi}{2}\cdot  \frac{\min{\left\{\left\| x_i - x_j \right\|_2, \delta + \eta(x) \right\}}}{\eta(x) + \delta}\right), \quad x \in \RR^{d \times n} \setminus{\Omega_{d \times n}}, \quad \delta > 0 \\
    \omega_{i, j}(x) &\coloneq \frac{1}{2}\frac{\tilde{\omega}_{i, j}(x)}{ \sum_{(l, k) \in H}{\tilde{\omega}_{l,k}(x)}}.
\end{split} \end{equation}
Here $\delta > 0$ is some fixed parameter. We note that for pairs $i,j$ for which $\|x_i-x_j\|$ is larger than the minimal distance $\eta(x)$ by $\delta$ or more, the weight $\tilde \omega_{i,j}(x)$ will be zero. Thus, the summation in the definition of $\Phi$ at worst includes all $n \choose 2 $ pairs, but will often include much less pairs. Smaller $\delta$ will lead to smaller support of $\Phi$, at the price of reduced stability. 

By construction, the weights $\omega_j$ are well defined for every $x\not \in \Omega_{d \times n}$, and they sum to $1/2$. These weights are defined to ensure continuity for $x$ approaching $\Omega_{d \times n}$, in the sense of \eqref{eq:cond_weak_convergence}. Informally, for $x$ close to $\Omega_{d \times n}$, the weights $\omega$ are then defined so that these near-identical columns dominate the other columns. Formally, we prove the following:

%It remains to show:
% to conclude the construction of $\mathcal{E}[\cdot]$. We now show that the operator is a universal, bounded, continuity-preserving and linear antisymmetric projection. Given lemma \ref{lemma:bec}, theorem \ref{theorem:universality_asc}, and lemma 4.2 at \cite{frames}, it is sufficient to show $\Phi[\cdot]$ defines a linear WS operator. As for the efficiency, $\mathcal{E}[f]$ requires as few as $\mathcal{O}\left(n^2 d \right)$ evaluations of any given MLP neural networks.
\begin{prop} \label{prop:stab_operator_correct}
Given the weighted permutation frame $\mathcal{F}_{\text{as}} $ defined at \eqref{eq:tau} and \eqref{eq:w}, the operator $\Phi[\cdot]$ defined in \eqref{def:staboperator} is a  \textit{WS} operator.
\end{prop}
\begin{proof}
	 We will show that all related conditions in the definition \ref{def:afa} hold for $\Phi$:
	 
	\textbf{$\Phi$ is Bounded} 
For every compact $K \subset \RR^{d \times n}$, which is $S_n$-invariant, and for any function $f: \RR^{d \times n} \to \RR$, we bound $\Phi[f]$ over $K$, by noting the for all $x \in  \Omega_{d\times n}$ we have $\Phi[f](x)=0 $, and for all $x \in \RR^{d\times n}\setminus \Omega_{d\times n}$,	
\begin{equation}
		\begin{split}
			\max_{x \in K}{\abs{\Phi[f](x)}} 
            &= 
			\max_{x \in K}{\abs{f(x) - \sum_{\sigma_{i,j} \in H}{\omega_{i,j}(x)\left(f(x) + f(\sigma_{i,j} x) \right)}}}
			\\
			&=
			\max_{x \in K}{\abs{\left(1 - \sum_{\sigma_{i,j} \in H}{\omega_{i, j}(x)}\right)f(x) - \sum_{\sigma_{i,j} \in H}{\omega_{i,j}(x) f(\sigma_{i,j} x)}}}
			\\
			&\leq 
			\frac{1}{2}\max_{x \in K}{\abs{f(x)}} + \sum_{\sigma_{i,j} \in H}{\omega_{i,j}(x) \max_{x \in K}{\abs{f(\sigma_{i,j} x)}}}
			\\
			&\leq \frac{1}{2}\max_{x \in K}{\abs{f(x)}} + \frac{1}{2}\max_{x \in K}{\abs{f(x)}} = \max_{x \in K}{\abs{f(x)}}
		\end{split}
	\end{equation} 
	\textbf{Antisymmetry}
	  Let $f:\RR^{d\times n}\to \RR$ be an antisymmetric function. Then, for all $x \in \RR^{d \times n}$, and for all $\sigma_{i,j} \in H$, we have $f(x) = -f(\sigma_{i,j} x)$, and so $\Phi[f] = f$. 
	
	\textbf{Continuity in $\RR^{d\times n}\setminus \Omega_{d\times n}$} The function $\eta: \RR^{d \times n} \to \RR$ is continuous everywhere, and so is the cotangent function over the domain $(0, \frac{\pi}{2}]$. Accordingly, for all $x \in \RR^{d \times n} \setminus{\Omega_{d \times n}}$, and for any fixed $\delta > 0$, we have $\frac{\pi\min{\left\{\left\| x_i - x_j \right\|_2, \delta + \eta(x) \right\}}}{2\eta(x) + 2\delta} \in (0, \frac{\pi}{2}]$ ensuring the continuity of $\tilde{\omega}_{i, j}$ for all $1 \leq i < j \leq n$. And so, for every $f\in  C\left(\RR^{d \times n} \setminus \Omega_{d\times n} \right)$, the function $\Phi[f]$ will also be continuous on $\RR^{d\times n}\setminus \Omega_{d\times n} $, as summing over $\sigma_{i, j} \in H$ the multiplication of the function $f(x)$, the functions $f(\sigma_{i,j}x)$ and the weight functions $\omega_{i,j}(x)$, which are all continuous outside of $\Omega_{d \times n}$, is continuous.
	
	\textbf{Continuous limit}
    Let $x^k \to x$ be any convergent sequence in $\RR^{d \times n} \setminus{\Omega_{d \times n}}$, with limit a $x \in \Omega_{d \times n}$ at which $\eta(x) = 0$. 
    To simplify, we define the complementary subsets $I^+ ,I^- \subseteq H$ by
    \begin{equation}
    \begin{split}
        I^+ &\coloneq \left\{i, j \in [n]: x_i = x_j, i \ne j \right\} \ne \emptyset, \\
        I^- &\coloneq H \setminus{I^+}.
    \end{split}
    \end{equation}
For all $(i,j)\in I^+$, we will have that $\|x_i^k-x_j^k\| \rightarrow 0 $, which will imply that $\tilde\omega_{i,j}(x^k)\rightarrow \infty $. In contrast, for every $(i, j) \in I^-$, the weight $ \tilde\omega_{i,j}(x^k) $ has a (finite) limit, and therefore, for every $(i, j) \in I^-$
\begin{equation}\label{eq:small_omega}
\begin{split}
    &\lim_{k \to \infty}{\omega_{i,j}(x^k)} \\  &= 
 \frac{1}{2} \lim_{k \to \infty}{
\frac{\tilde{\omega}_{i,j}(x^k)}{\sum_{(l, k) \in I^+}{\tilde{\omega}_{l,k}(x^k)}  + \sum_{(l, k) \in I^-}{\tilde{\omega}_{l,k}(x^k)}
} 
} \\ &\leq
 \frac{1}{2} \lim_{k \to \infty}{
\frac{\tilde{\omega}_{i,j}(x^k)}{\sum_{(l, k) \in I^+}{\tilde{\omega}_{l,k}(x^k)}
} 
} = 0.
\end{split}
\end{equation}
Now, let $f \in C(\RR^{d \times n})$ be any continuous function. By definition, the sequence $\abs{f(x^k) - f(\sigma_{i, j} x^k)}$ converges to zero for all $(i, j) \in I^+$ for any such function $f$. Finally, considering \eqref{eq:small_omega}, we hereby conclude the continuity of $\Phi[\cdot]$ with the limit
\begin{equation}
\begin{split}
     &\lim_{k \to \infty}{\left(
f(x^k) - \sum_{\sigma_{i,j} \in H}{\omega_{i,j}(x^k)\left(f(x^k) + f(\sigma_{i, j} x^k) \right)}
\right)} \\ &= 
\lim_{k \to \infty}{\left(
f(x^k) - \sum_{\sigma_{i,j} \in I^+}{\omega_{i,j}(x^k)\left(f(x^k) + f(\sigma_{i, j} x^k) \right)}
\right)}
\\ &= 
\lim_{k \to \infty}{\left(
f(x^k) - 2\sum_{\sigma_{i,j} \in I^+}{\omega_{i,j}(x^k)f(x^k)}
\right)} \\
&= \lim_{k \to \infty}{\left(f(x^k) - f(x^k) \right)} = 0
\end{split}
\end{equation}
as desired, to show that $\Phi[\cdot]$ is a linear WS operator.
\end{proof}
% \begin{remark}
%     In \href{sec:appendixB}{Appendix B} we discuss an extension of the method to nonlinear projection operators.
% \end{remark}
\textbf{Complexity} In current terms, our antisymmetric frame averages over $\mathcal{O}(n^3)$ permutations: the $\mathcal{O}(n)$ permutations determined by the frame $\mathcal{F}_{as}$ \eqref{eq:tau}, composed with all $\mathcal{O}(n^2)$ transpositions from the set $H$ \eqref{eq:all_transpositions}. By the definition of our frame, each transposition $\sigma_{i,j}\in H$ is weighted by a non-negative value $\omega_{i, j}(x)$, which is proportional to the input $x \in \RR^{d \times n} \setminus \Omega_{d \times n}$ and some constant $\delta > 0$, as defined at \eqref{def:small_omega}. In fact, these weights are permutation invariant, thus can be pre-computed for any given $x \in \RR^{d \times n} \setminus \Omega_{d \times n}$. Accordingly, whenever $\omega_{i, j}(x) = 0$, we can preemptively exclude $\sigma_{i, j}$ from the summation in \eqref{def:staboperator}, which raises the question: given a small enough $\delta > 0$, how often can we ignore elements of $H$? The answer to that question lies in the largest number of repeated shortest distances
among $n$ points in $\RR^{d \times n} \setminus \Omega_{d \times n}$, and has already been answered; using the result of H. Harborth \cite{Harborth} for $d=2$, and later proven by K. Bezdek \cite{BEZDEK2002192} for $d\geq 3$, we in fact use only $\mathcal{O}(n^2)$ permutations in our frame overall. 

\section{Experiments}
To experimentally evaluate the efficiency of our bi-Lipschitz continuous ansatz, and the weighted frame ansatz, we compare them against the Vandermonde determinant \eqref{eq:vandermonde} based approach from \cite{ye2024widetilde} and a naive Multi-Layer-Perceptron (MLP) estimator which does not enforce antisymmetry. We use these different models to address a simple antisymmetric regression task: approximating the antisymmetric function $f(A) = \det{A}: \mathbb{R}^{n \times n} \rightarrow \RR$.  The experiments code is available on GitHub \footnote{\url{https://github.com/Matt88Ac/blansatz}}.

\subsection{Setup} \label{exp_setup}
We generated three training sets of 110,000 matrices of orders $n \in \{10, 15, 20\}$, with a corresponding validation set of 15,000 matrices and a test set of 20,000 matrices, all uniformly sampled from $\left[0, 1.1\right]^{n \times n}$, then scaled so that the determinants approximately distribute normally with a non-zero mean and higher variance. We moreover train models that are antisymmetric by design exclusively on positive determinants by applying a random transposition to the input whenever the output should have been negative. This benefits the training process by optimizing the learned parameters over a compressed range of determinants, with a larger empirical mean, allowing faster convergence and improved generalization. Of course, the fairness in performance comparison is retained; Due to their inherited equivariance, antisymmetric models in fact learn the determinant of each matrix of the training set up to any permutation simultaneously, hence the training process remains equivalent across all models. We optimize the parameters with respect to the $\|\cdot\|_{l_1}$ metric using ADAM, with the same learning rate and learning rate scheduler settings across all methods. We implemented the symmetric functions required for the ansatz \cite{ye2024widetilde} using the Deep Sets model \cite{zaheer2017deep}, and sampled $n^2 + 1$ constant vectors from $S^{n-1}$ for the Vandermonde determinant computation.

\begin{table}[h]
\centering
\begin{tabular}{lrrrrrr}
\toprule
 &$n = 10$ & $n = 15$ & $n = 20$ \\
Ansatz &  &  &  &  &  &  \\
\midrule
Antisymmetric Weighted Frame (ours) & 63,503 & 141,766 & 193,301 \\
Bi-Lipschitz (ours) & 79,741 & 121,501 & 275,121 \\
MLP & 223,489 & 783,593 & 993,793 \\
\textit{YLGLHW} \cite{ye2024widetilde} & 704,677 & 1,649,122 & 3,054,417\\
\bottomrule
\end{tabular}
\caption{Ansatz's number of parameters}
 \label{exp:params}
\end{table}
For the MLP method, we used the same depth as the MLP used within our construction \ref{eq:pi}, with a width similar to, or wider than, that of our hidden layers. %In our ansatz, we construct $2n^2 + 1$ parametric functions, as described in \ref{thm:multi}, to handle the input, which is why our model uses more parameters in a case where the hidden layers of the MLP have the same shape. 
Note that the MLP ansatz does not respect the antisymmetry.
The number of parameters in each model is summarized in Table \ref{exp:params}.
 
\subsection{Results}
\begin{figure}[h]
\begin{subfigure}{0.33\linewidth}
\includegraphics[width=1\linewidth, height=2.85cm]{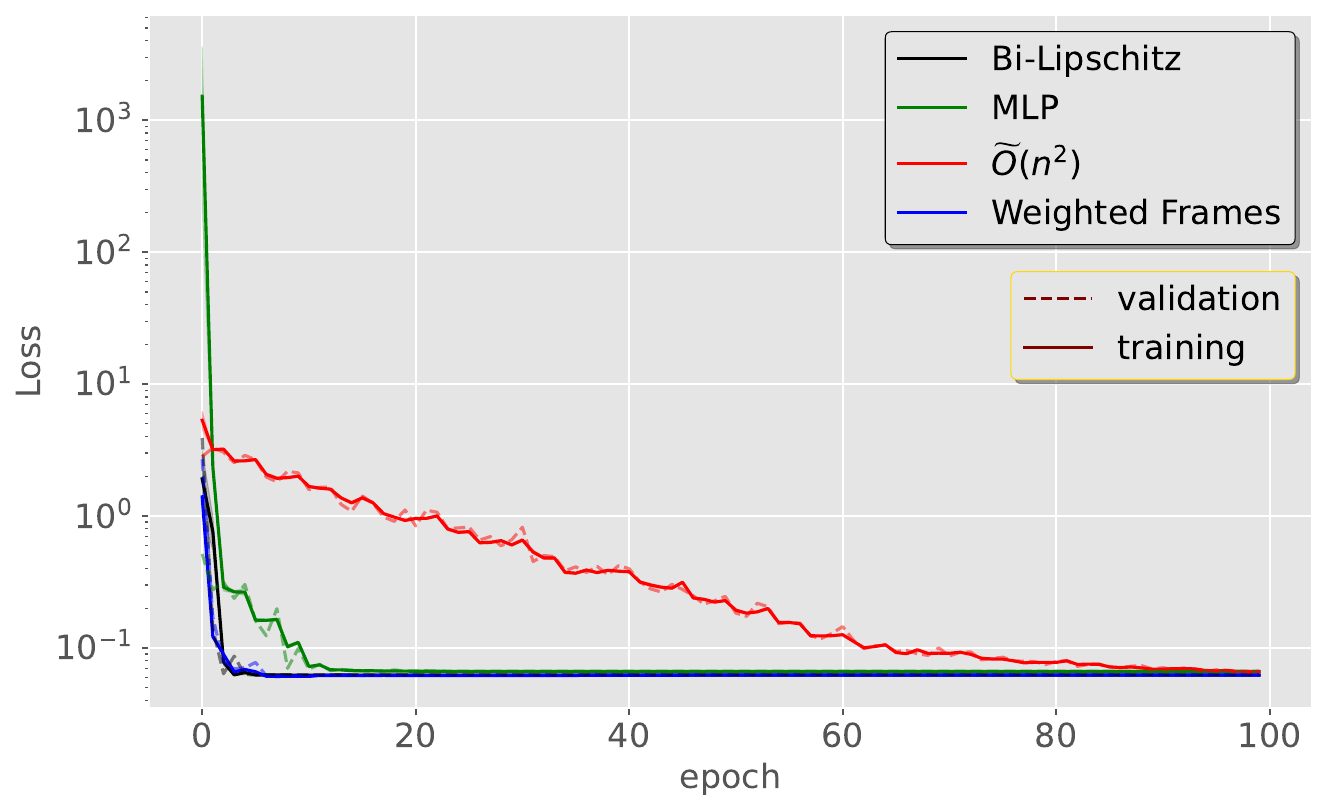} 
\caption{$n = 10$}
\label{fig:det10}
\end{subfigure}
\begin{subfigure}{0.33\linewidth}
\includegraphics[width=1\linewidth, height=2.85cm]{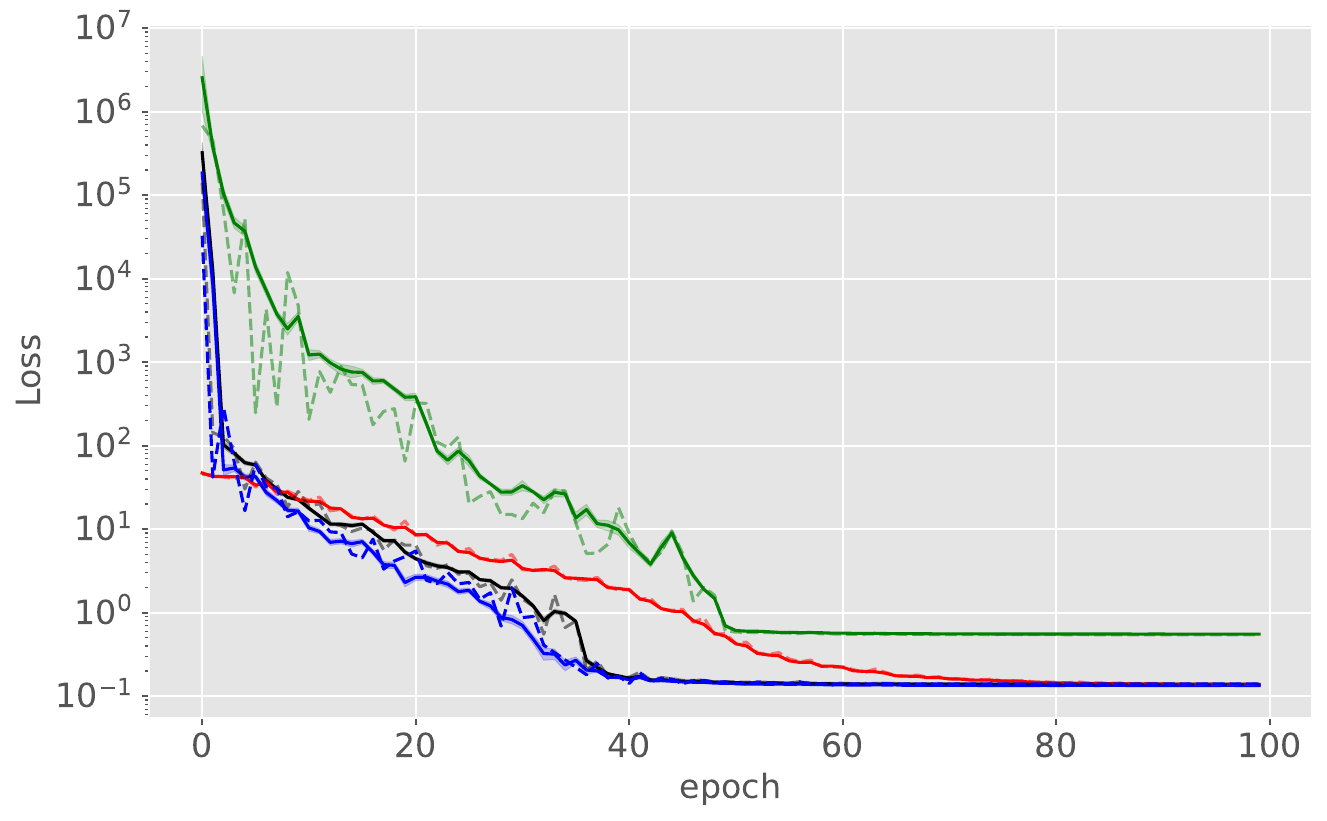}
\caption{$n = 15$}
\label{fig:det15}
\end{subfigure}
\begin{subfigure}{0.33\linewidth}
\includegraphics[width=1\linewidth, height=2.85cm]{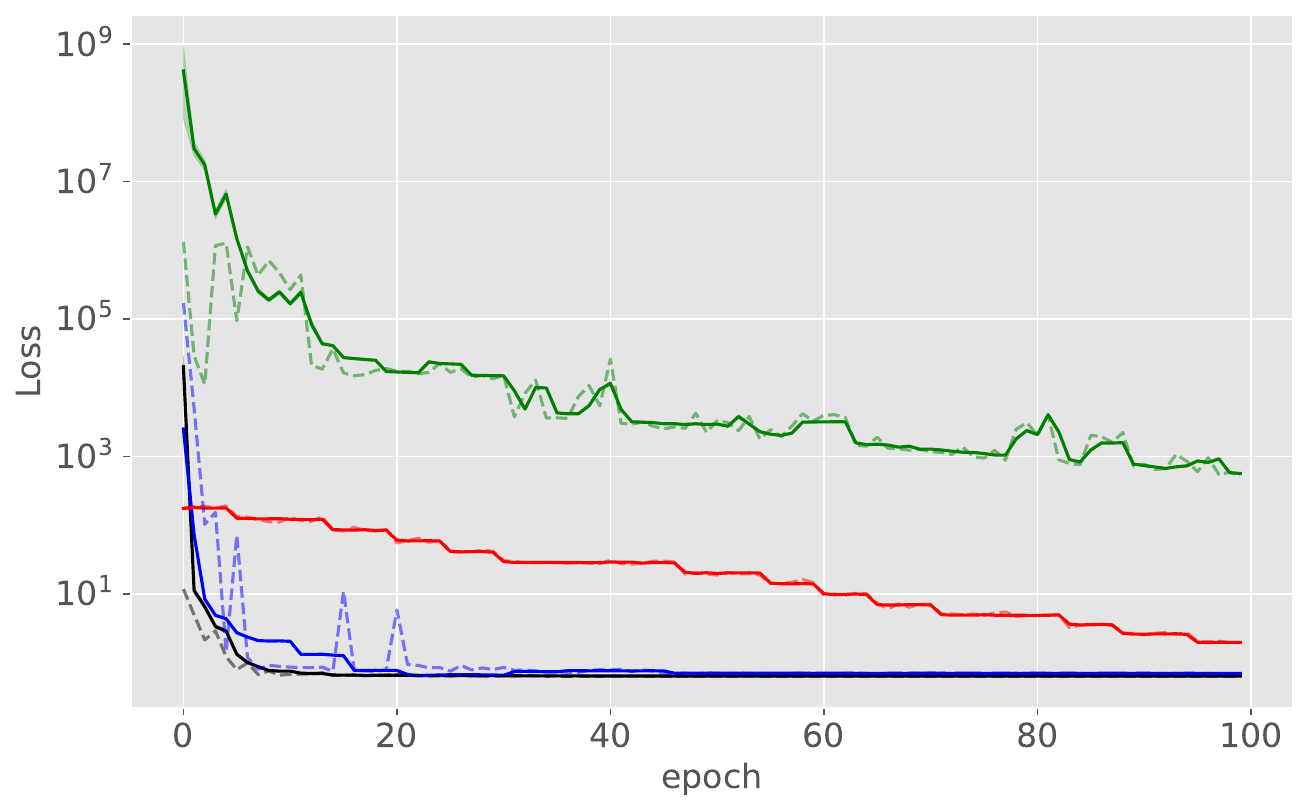}
\caption{$n = 20$}
\label{fig:det20}
\end{subfigure}
\caption{Training-Validation convergence comparison of our ansatz (black), against \cite{ye2024widetilde}'s (red) and an MLP (green), for approximating the determinant of matrices of different orders (denoted by $n$). Our ansatz exhibits faster convergence and better or comparable accuracy. }
\label{fig:exp_det_train_val}
\end{figure}

The accuracy of the different models is shown in Table \ref{exp:res} in terms of the Mean Absolute Error (which was used for training).
% and the Mean Absolute Relative Error. 
In Figure \ref{fig:exp_det_train_val}  we show the convergence rates of the different methods for this experiment. For all values of $n$ tried, our models were more accurate than the competing models, sometimes by a large margin. Additionally, the convergence of our methods was much faster. The results also show that for $n=15,20$ the performance of the MLP is substantially lower than that of other antisymmetric models, indicating the importance of imposing the antisymmetric structure.

\begin{table}[!h]
\centering
\begin{tabular}{lrrrrrr}
\toprule
 & \multicolumn{2}{c}{$n = 10$} & \multicolumn{2}{c}{$n = 15$} & \multicolumn{2}{c}{$n = 20$} \\
 & MAE & MARE & MAE & MARE & MAE & MARE \\
Ansatz &   &  &  \\
\midrule
A.S. Weighted Frames (ours) & \bfseries 0.0582 & \bfseries 0.00048 & \bfseries 0.1384 & \bfseries 0.000509 & 0.7781 & 0.00055 \\
Bi-Lipschitz (ours) & \color{black} 0.0633 & 0.0005 & \color{black} 0.1400 & 0.000515 & \color{black} \bfseries 0.6626 & \bfseries 0.0005 \\
MLP & 0.0676 & 0.000534 & 0.5769 & 0.002123 & 538.83 & 0.4079 \\
\textit{YLGLHW} \cite{ye2024widetilde} & 0.0666 & 0.000523 & 0.1421 & 0.000526 & 2.0044 & 0.0015 \\
\bottomrule
\end{tabular}
\caption{Mean Absolute Error (MAE) and Mean Absolute Relative Error (MARE) evaluation of the ansatzs, on the test sets, with $n$ being the order of the matrices evaluated. Best of each measurement is in bold.}
  \label{exp:res}
\end{table}

\newpage
\section{Conclusions}
This paper addresses the problem of constructing continuous, universal, and computationally efficient neural network ansatzes for antisymmetric functions, motivated by applications in quantum many-body systems and fermionic wave function approximation. While neural networks provide powerful approximation capabilities, enforcing antisymmetry in a stable and principled manner remains challenging, and existing approaches often suffer from instability, discontinuity, or exponential complexity.
We introduced two frameworks for antisymmetric function approximation.

The first framework is based on $A_n$-invariant bi-Lipschitz embeddings with respect to a metric on the quotient space $\RR^{d \times n} / A_n$, which is natural to the problem. By embedding the quotient space into a Euclidean space in a bi-Lipschitz manner, antisymmetric function approximation is reduced to standard function approximation followed by a simple antisymmetrization step. This construction yields ansatzes that are antisymmetric and continuous by construction, and enables quantitative approximation guarantees.

The second framework is based on antisymmetric frame averaging networks, which generalize recent permutation averaging techniques while avoiding full group symmetrization. We constructed a bounded, continuity-preserving antisymmetric projection operator using weighted permutation frames of polynomial cardinality, together with a weakly-stabilizing operator. This approach produces a broad family of antisymmetric ansatzes that are efficient to compute, continuous, and universal, while requiring only $\mathcal{O}(n^3)$ permutations (at the worst case) rather than factorial complexity. We established universality results for this framework and showed that it achieves approximation rates comparable to those of standard neural networks. Finally, we complemented the theoretical results with preliminary numerical experiments on learning determinants of matrices of varying sizes. Both proposed ansatzes demonstrated faster convergence and improved or comparable accuracy compared with existing antisymmetric constructions and naive MLP baselines, highlighting the importance of explicitly enforcing antisymmetry, especially in higher dimensions.

\section{Appendix A: Quantitative Bounds for $A_n$ Invariant Embedding} \label{sec:appendixA}
In this section, we quantify the Lipschitz bounds of our $A_n$-invariant embedding defined at \eqref{eq:xi}, for all $n, d \in \NN$, and some $m \in \NN$, $A \in \RR^{m \times d}$, with respect to the semi metric on $\RR^{d \times n}$ 
\begin{equation}
    d_{p_{+}}(x, y) \coloneq \min_{\sigma \in A_n}{\left\|x - \sigma y \right\|_p}, \quad p \geq 1.
\end{equation}
We will begin with the case $d=1$, and then move on to the case $d>1$.
\subsection{Bi-Lipschitz bounds for $d=1$}
In the $d=1$ case, we consider the mapping 
$$\xi(x) = \begin{bmatrix}
\sort{(x)} \\ Q(x)
\end{bmatrix}$$
from the main text. We will prove that this mapping is bi-Lipschitz, with a moderate constant that does not depend on $n$:
\begin{theorem}\label{thm:quantitative}
For every $n\in \NN$ and $p\geq 1$, 
\begin{equation}
   2^{\frac{1-p}{p}} d_{p_{+}}(x, y) \leq \|\xi(x)-\xi(y)\|_p \leq (1+2^{p-1})^{1/p}  d_{p_{+}}(x, y) 
\end{equation}
\end{theorem}
Throughout the proof, we will use the fact that 
\begin{equation}
    \|\xi(x)-\xi(y)\|_p^p=\|\sort(x)-\sort(y)\|_p^p+|Q(x)-Q(y)|^p=d^p_{p_\pm}(x, y) +|Q(x)-Q(y)|^p
\end{equation}
We will prove the lower bound in Theorem \ref{thm:quantitative} by proving the following lemma:
\begin{lemma} \label{lemma:l1}
    The inequality
    \begin{equation}
       2^{1-p}d^p_{p_+}(x, y) \leq d^p_{p_\pm}(x, y) +  \abs{Q(x)- Q(y)}^p 
    \end{equation}
     holds for all $x, y \in \RR^{n}$, and $p \geq 1$.
\end{lemma}
\begin{proof}
    Let $x, y \in \RR^n$ be any two vectors. For each, we denote 
        \begin{equation} \begin{split}
        \sigma_x &\coloneq \argsort{\left(x \right)}, \quad \hat{x} \coloneq \sigma_x x, \quad i \coloneq \argmin_{1 \leq i < n}{\abs{\hat{x}_{i+1} - \hat{x}_{i}}} \\
        \sigma_y &\coloneq \argsort{\left(y \right)}, \quad \hat{y} \coloneq \sigma_y y,  \quad j \coloneq \argmin_{1 \leq j < n}{\abs{\hat{y}_{j+1} - \hat{y}_{j}}}
    \end{split} \end{equation}
and define the transpositions $\tau_i = (i \quad i+1)$ and $\tau_j = (j \quad j+1)$ in $S_n$. 
When $\sigma_x$ and $\sigma_y$ have the same sign, the inequality holds trivially because $d^p_{p_+}(x, y)=d^p_{p_\pm}(x, y) $ . Otherwise, considering that for any transposition $\tau \in S_n$, the inequality
\begin{equation}
    \min_{\sigma \in A_n}{\left\| x - \sigma y\right\|_p} \leq \left\| \hat{x} - \tau \hat{y}\right\|_p 
\end{equation}
holds, we can use the Minkowski inequality, so that for all $p \geq 1$, we have
\begin{equation}
    \begin{split}
        2d_{p_+}(x, y) &= 
2 \min_{\sigma \in A_n}{\left\| x - \sigma y\right\|_p}   \\
&\leq \left\| \hat{x} - \tau_j\hat{y}\right\|_p + \left\| \tau_i\hat{x} - \hat{y} \right\|_p \\
&= \left\| \hat{x} - \tau_j\hat{y} + \hat{y} - \hat{y}\right\|_p + \left\| \tau_i\hat{x} - \hat{y} + \hat{x} - \hat{x}\right\|_p^p  \\
&\leq 2\|\hat x-\hat y\|_p + \left\|\hat{y} - \tau_j\hat{y} \right\|_p + \left\|\hat{x} - \tau_i\hat{x} \right\|_p  \\
&= 2d_{p_{\pm}}(x, y) + 2\abs{Q(x)} + 2\abs{Q(y)}\\
&=2d_{p_{\pm}}(x, y) + 2\abs{Q(x)-Q(y)}
    \end{split}
\end{equation}
where for the last equality we used the fact that the sign of $Q(x)$ is the sign of $\tau_x$. Thus, using the convexity of $\abs{\cdot}^p$ and Jensen's inequality, we have,
\begin{equation}
    \begin{split}
        d_{p_+}^p(x, y) &\leq \abs{d_{p_{\pm}}(x, y) + \abs{Q(x)- Q(y)}}^p \\
        & = 2^p \abs{\frac{1}{2}d_{p_{\pm}}(x, y) + \frac{1}{2}\abs{Q(x)-  Q(y)}}^p  \\
        &\leq 2^{p-1}\left(d^p_{p_{\pm}}(x, y) + \abs{Q(x)-Q(y)}^p \right) \\
    \end{split}
\end{equation}
multiplying this inequality by $2^{1-p}$ gives the desired result.
\end{proof}
We will now prove the upper bound in Theorem \ref{thm:quantitative} by proving the following lemma:

\begin{lemma} \label{lemma:lemma_Q}
    For all $x, y \in \RR^n$  and $p \geq 1$,
    \begin{equation}
        \abs{Q(x) - Q(y)}^p \leq 2^{p - 1}d^p_{p_+}(x, y).
    \end{equation}
\end{lemma}
The bound in the theorem is easily obtained from the lemma by adding this inequality to the trivial inequality  $d^p_{p_\pm}(x, y)\leq d^p_{p_+}(x, y) $.
\begin{proof}[Proof of Lemma \ref{lemma:lemma_Q} ]
    For all $x, y \in \RR^{n}$ we define as before
    \begin{equation} \begin{split}
        \sigma_x &\coloneq \argsort{\left(x \right)}, \quad \hat{x} \coloneq \sigma_x x \\
        \sigma_y &\coloneq \argsort{\left(y \right)}, \quad \hat{y} \coloneq \sigma_y y
    \end{split} \end{equation}
    So, we have
    $$\abs{Q\left(x \right) - Q\left(y \right)} = 
    \abs{(-1)^{\sigma_x} Q(\hat{x}) - (-1)^{\sigma_y}Q(\hat{y})}.
    $$
    Assume without a loss of generality that $Q(\hat{x}) \geq Q(\hat{y})$. We now divide into two cases:
    \begin{enumerate}
        \item Case 1: $(-1)^{\sigma_x} = (-1)^{\sigma_y}$. Then, there exists $s \in [n-1]$ such that 
        \begin{equation}
            \begin{split}
                \abs{Q(x) - Q(y)}^p &= \abs{\min_{1 \leq i < n}{\left(\hat{x}_{i+1} - \hat{x}_i\right)} - \left(\hat{y}_{s+1} - \hat{y}_{s}\right) }^p  \\
                &\leq \abs{\left(\hat{x}_{s+1} - \hat{x}_s\right) - \left(\hat{y}_{s+1} - \hat{y}_s\right)}^p  \\
                &\leq 
                2^{p - 1}\left(\abs{\hat{x}_{s+1} - \hat{y}_{s+1}}^p + \abs{\hat{x}_s - \hat{y}_s}^p \right)  \\
                &\leq 
                2^{p - 1}\sum_{i \in [n]}{\abs{\hat{x}_i - \hat{y}_{i}}}^p  \\
                &\leq 2^{p - 1} d^p_{p_+}(x, y).
            \end{split}
        \end{equation}
        \item Case 2: $(-1)^{\sigma_x} \ne (-1)^{\sigma_y}$. Then, there exists $s \in [n-1]$, and a transposition $\tau \in S_n$, such that
        \begin{equation}
        \begin{split}
            \abs{Q(x) - Q(y)}^p &= \abs{\left(\hat{x}_{s+1} - \hat{x}_s\right) + \min_{1 \leq i < n}{\left(\hat{y}_{i+1} - \hat{y}_i\right)}}^p  \\
            &\leq  \abs{\left(\hat{x}_{s+1} - \hat{x}_s\right) - \left(\hat{y}_{\tau(s+1)} - \hat{y}_{\tau(s)}\right)}^p \\
            &\leq 2^{p - 1}\left(\abs{\hat{x}_{s+1} - \hat{y}_{\tau(s+1)}}^p + \abs{\hat{x}_s - \hat{y}_{\tau(s)}}^p \right)  \\
            &\leq 2^{p - 1}\sum_{i \in [n]}{\abs{\hat{x}_i - \hat{y}_{\tau(i)}}^p}  
 \\ &\leq 2^{p - 1}d^p_{p_+}(x, y).
        \end{split}
        \end{equation}
    \end{enumerate}
And so, we have the inequality
$$    \abs{Q(x) - Q(y)}^p \leq 2^{p - 1}d^p_{p_+}(x, y)
$$
holds for all $x, y \in \RR^n$, to conclude the proof of the lemma. 
\end{proof}

\subsection{Bi-Lipschitz bounds for $d>1$}
We now consider Bi-Lipschitz bounds for the case $d>1$.  Namely, for  $n, d ,m\in \NN$, $p \geq 1$, and $A \in \RR^{m \times d}$, we find constants
\begin{equation} \begin{split}
    c &\coloneq c(n, p, A) \\
    C &\coloneq C(n, p, A)
\end{split} \end{equation}
such that
\begin{equation}
    c d_{p_+}(x, y) \leq \left\|\xi\left(x ; A \right) - \xi\left(y ; A \right) \right\|_{p} \leq Cd_{p_+}(x, y).
\end{equation}
We will begin with the upper-Lipschitz bounds. These bounds are formulated in terms of the $p$-operator norm of $A$, namely
\begin{equation}
   \|A\|_{\text{op},p}=\max_{x: \|x\|_p=1} \|Ax\|_p  
\end{equation}
\begin{theorem} \label{theorem:F_a_bl_upper}
    Let $n, d$ and $m$ be natural numbers, and $p \geq 1$ a real number. Then, for any matrix $A \in \RR^{m \times d}$, we have   for all $x, y \in \RR^{d \times n}$ that
    \begin{equation}
        \|\xi\left(x ; A \right) - \xi\left(y ; A \right)\|_p \leq C \cdot d_{p_+}(x, y), \text{ where }  C = \left(2^{p-1} + 1\right)^{\frac{1}{p}}  \cdot \|A\|_{\text{op},p}
    \end{equation}
    \end{theorem}

\begin{proof}
For given $x,y\in \RR^{d\times n}$, let $\sigma_\ast$ the permutation in $A_n$ for which 
\begin{equation}
   d_{p_+}(x,y)=\|x-\sigma_\ast y\|_p . 
\end{equation}
Then we have 
\begin{equation}
    \begin{split}
        \|\xi\left(x ; A \right) - \xi\left(y ; A \right)\|^p_p &=   \sum_{k\in [m]} \|\xi(x^Ta_k)-\xi(y^Ta_k) \|_p^p \\
        &\stackrel{\text{Theorem \ref{thm:quantitative}}}{\leq} \left(2^{p - 1} + 1 \right)  \sum_{k \in [m]}{d^p_{p_+}(x^Ta_k, y^Ta_k)}  \\
        &\leq \left(2^{p - 1} + 1 \right)  \sum_{k \in [m]}{\sum_{i \in [n]}{\left\|a_k^T(x_i-y_{\sigma_\ast(i)}) \right\|_p^p}}  \\
        &= \left(2^{p - 1} + 1 \right) \sum_{i \in [n]}{\left\|A\left( x_i - y_{\sigma_\ast(i)} \right) \right\|^p_p}  \\
&\leq \left(2^{p - 1} + 1 \right) \cdot \|A\|_{\text{op},p}^p
\sum_{i \in [n]}{\left\|\left( x_i - y_{\sigma_\ast(i)} \right) \right\|^p_2}  \\
& \leq \left(2^{p - 1} + 1 \right) \cdot  \|A\|_{\text{op},p}^p \cdot d^p_{p_+}(x, y)
\end{split}
\end{equation}
\end{proof}
Proving a lower-Lipschitz bound requires some assumptions on $A$ (in particular, that the mapping defined by $A$ is injective, up to $A_n$ symmetries). We will focus on matrices $A \in \RR^{m \times d}$ with the {projective uniformity (PU)} property: for $p\geq 1, l \in [m]$, and $\delta > 0$,  we will say that $A$  if ($l$, $\delta$, $p$)-PU if for all $x \in \RR^d$ with $\|x\|_p = 1$, the $l$th smallest value of $\left(\abs{a_k^T x} \right)_{k \in [m]}$ exceeds $\delta$. Projective uniformity is a notion defined in \cites{Cahill2025, doi:10.1137/12089939X}, and it was used in \cite{dym2025quantitativeboundssortingbasedpermutationinvariant} to compute lower Lipschitz constants for the $S_n$-invariant function $\beta_A$ in the case $p=2$. Here we use similar techniques to address $A_n$-invariant functions, in the more general $p\geq 1$ setting. We now state the main theorem of this section:
\begin{theorem} \label{theorem:F_a_bl_lower}
    Let $n, d, l$ and $m \geq n^2(l-1)$ be natural numbers, and $p \geq 1$, $\delta > 0$ be real numbers. Then, for all ($l$, $\delta$, $p$)-PU exhibiting matrix $A \in \RR^{m \times d}$, we have for all $x, y \in \RR^{d \times n}$ that 
    \begin{equation}
        c \cdot d_{p_+}(x, y) \leq \|\xi\left(x ; A \right) - \xi\left(y ; A \right)\|_p, \text{ where }  c =\delta2^{\frac{1-p}{p}} \left(m-n^2(l-1))\right)^{1/p}.
    \end{equation}
\end{theorem}
\begin{proof}
Let $x, y \in \RR^{d \times n}$ be arbitrary matrices, and let $A \in \RR^{m \times d}$ be any matrix that exhibits ($l$, $\delta$, $p$)-PU, given the real numbers $p \geq 1, \delta > 0$, and the natural numbers $l \in \NN$ and $m \geq n^2(l-1)$.  We now begin our proof with the lower Lipschitz bound. We first note that 
\begin{equation}\label{eq:to1d}
\|\xi\left(x ; A \right) - \xi\left(y ; A \right)\|_p^p=\sum_{k=1}^m \|\xi(x^Ta_k) - \xi(y^Ta_k)\|_p^p \stackrel{\text{ Lemma \ref{lemma:l1}}}{\geq} 2^{1-p} \sum_{k=1}^m d_{p_+}^p(x^Ta_k,y^Ta_k)
\end{equation}
and therefore it is sufficient to bound the expression on the right-hand side. To do so, for fixed $x,y$ and fixed indices $i,j\in [n]$, let $S_{i,j}$ denote the set 
\begin{equation}
   S_{i,j}=\left\{k\in [m] :  \abs{a_k^T(x_i-y_j)}_p \geq \delta \|x_i-y_j\|_p \right\} 
\end{equation}
By the projective uniformity assumption, the complement of this set is at most of cardinality $l-1$. This implies that the complement of $S=\cap_{i,j\in [n]}S_{i,j} $ is of cardinality of at most $n^2(l-1)$, or in other words 
\begin{equation}
  |S|\geq m-n^2(l-1).   
\end{equation}
It follows that 
\begin{equation} \begin{split}\sum_{k=1}^m d_{p_+}^p(x^Ta_k,y^Ta_k)&\geq \sum_{k\in S} d_{p_+}^p(x^Ta_k,y^Ta_k)\\
&= \sum_{k\in S} {\min_{\sigma \in A_n}{
\sum_{i=1}^{n}{
\abs{a_k^T\left(x_i - y_{\sigma(i)} \right)}^p}}} \\
&\geq  \sum_{k\in S} {\min_{\sigma \in A_n}{
\sum_{i=1}^{n}{
\delta^p \|x_i-y_{\sigma(i)}\|_p^p}}}\\
&=\delta^p|S|d_{p_+}^p(x,y)\\
&\geq \delta^p(m-n^2(l-1))\cdot d_{p_+}^p(x,y)
\end{split} \end{equation}
Combining this with \eqref{eq:to1d} we obtain the desired result 
\begin{equation}
   \|\xi\left(x ; A \right) - \xi\left(y ; A \right)\|_p^p\geq 2^{1-p} \delta^p(m-n^2(l-1))\cdot d_{p_+}^p(x,y) 
\end{equation}
\end{proof}
\textbf{Projective uniform constructions} Our lower Lipschitz bounds are quantified in terms of projective uniformity. We can get more explicit bounds by considering explicit choices of $A$. For example, \cite{dym2025quantitativeboundssortingbasedpermutationinvariant} consider the case $p=2,d=2$, and showed that if $A$ is chosen to have  $m=4n^2$ rows which are chosen in uniform spaces in the unit circle, then $A$ will be $(l,\delta,p)$-PU for $l=3,\delta=\frac{1}{2n^2}$. The lower bound in this case will be 
\begin{equation}
   c =\delta2^{\frac{1-p}{p}} \left(m-n^2(l-1))\right)^{1/p}=\frac{1}{2n^2}2^{-1/2}(2n^2)^{1/2}=\frac{1}{2n}. 
\end{equation}
They also show that the $2$-operator norm of $A$ in this case is 
$\|A\|_{\text{op},2}=\sqrt{2} n$, and thus 
\begin{equation}
   C = \left(2^{p-1} + 1\right)^{\frac{1}{p}}  \cdot \|A\|_{\text{op},p}=\sqrt{3}\cdot \sqrt{2} \cdot n   
\end{equation}
and so the distortion of $\xi$, which is the ratio $C/c $, is at worst $\frac{\sqrt 3}{\sqrt 2}n^2 $, which is quadratic in $n$. This is very similar to the constants obtained in  \cite{dym2025quantitativeboundssortingbasedpermutationinvariant} for the $S_n$ invariant function $\beta_A$. Indeed, the lower and upper Lipschitz bounds we obtained in our theorems for $\xi$ (for the case $p=2$) differ only by a small constant factor from the lower and upper Lipschitz bounds obtained by \cite{dym2025quantitativeboundssortingbasedpermutationinvariant} for $\beta_A$.  In particular, that paper showed that for $d>2$, when considering randomly generated $A$, with high probability the distortion of $\beta_A$ will be $\sim n^2 $ when $m \sim n^2$ rows are considered (when ignoring logarithmic factors). These results carry over to our $\xi$ as well.

\newpage
\bibliographystyle{amsxport}
\bibliography{antisymmetry}

\end{document}